\documentclass[letterpaper]{article} 
\usepackage{aaai2026}  
\usepackage{times}  
\usepackage{helvet}  
\usepackage{courier}  
\usepackage[hyphens]{url}  
\usepackage{graphicx} 
\usepackage{natbib}  
\usepackage{caption} 
\usepackage{algorithm}
\usepackage{algorithmic}

\usepackage{newfloat}
\usepackage{listings}
\DeclareCaptionStyle{ruled}{labelfont=normalfont,labelsep=colon,strut=off} 
\floatstyle{ruled}
\newfloat{listing}{tb}{lst}{}
\floatname{listing}{Listing}
\usepackage{bibentry}

\title{PanoNav: Mapless Zero-Shot Object Navigation with Panoramic Scene Parsing and Dynamic Memory}
\author {
    Qunchao Jin,
    Yilin Wu,
    Changhao Chen*
}
\affiliations {
    PEAK-Lab, The Hong Kong University of Science and Technology (Guangzhou)\\
    qunchao.jin@outlook.com, ywu044@connect.hkust-gz.edu.cn, *Corresponding Author: changhaochen@hkust-gz.edu.cn
}

\usepackage{makecell}
\usepackage{pifont}
\usepackage{booktabs}
\usepackage{amsmath}
\usepackage{multirow}
\usepackage{morefloats}
\usepackage{pifont}

\begin{document}

\maketitle

\begin{abstract}

Zero-shot object navigation (ZSON) in unseen environments remains a challenging problem for household robots, requiring strong perceptual understanding and decision-making capabilities. While recent methods leverage metric maps and Large Language Models (LLMs), they often depend on depth sensors or prebuilt maps, limiting the spatial reasoning ability of Multimodal Large Language Models (MLLMs). Mapless ZSON approaches have emerged to address this, but they typically make short-sighted decisions, leading to local deadlocks due to a lack of historical context. We propose PanoNav, a fully RGB-only, mapless ZSON framework that integrates a Panoramic Scene Parsing module to unlock the spatial parsing potential of MLLMs from panoramic RGB inputs, and a Memory-guided Decision-Making mechanism enhanced by a Dynamic Bounded Memory Queue to incorporate exploration history and avoid local deadlocks. Experiments on the public navigation benchmark show that PanoNav significantly outperforms representative baselines in both SR and SPL metrics.
\end{abstract}
\begin{figure*}[t]
\centering
\includegraphics[width=\linewidth]{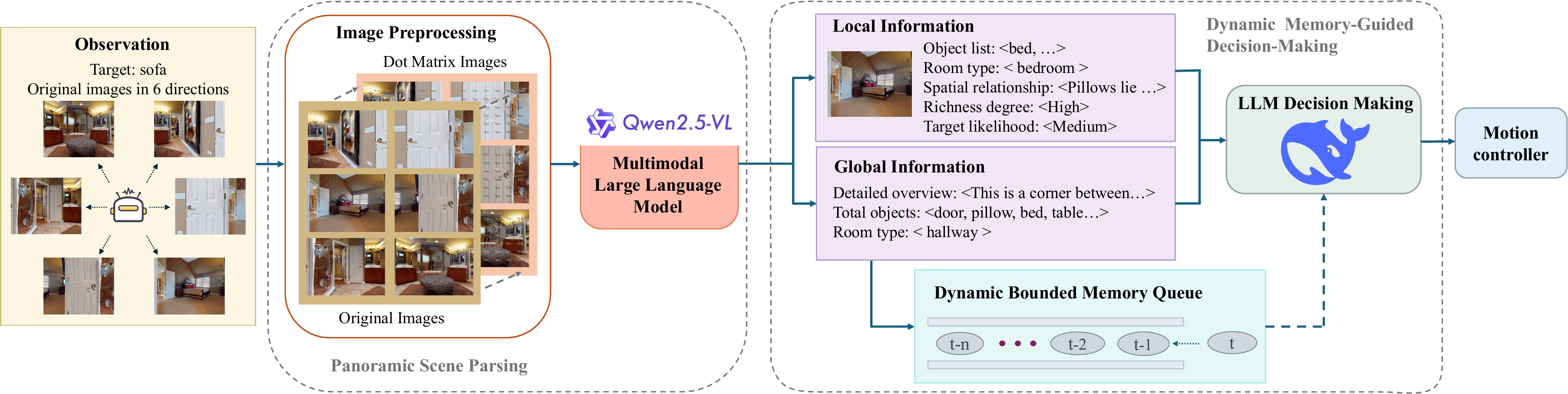}
\caption{Overview of the proposed PanoNav framework. At each timestep, the robot captures six directional RGB images to form a panoramic view. Each image is preprocessed into a dot matrix map, and both the RGB and dot matrix images are fed into an MLLM for spatial parsing. The model outputs local directional descriptions and a global scene summary. Global summaries are stored in a Dynamic Bounded Memory Queue, and together with local and global information, are used by the LLM to make navigation decisions that guide the robot’s movement.}
\label{fig:overview}
\end{figure*}

\section{Introduction}

Object Goal Navigation (ObjectNav)~\cite{sun2024survey} is a fundamental capability for domestic robots to effectively fulfill their functions. This task requires a robot to identify a specific object described by a human, plan a path toward it, and actively navigate through the environment to reach the target. As a prerequisite to downstream tasks—such as robotic search and exploration, mobile manipulation~\cite{billard2019trends} and human-robot interaction~\cite{ajoudani2018progress}—reliable ObjectNav is critical for enabling autonomous robotic assistance.

Since the introduction of standardized definitions and evaluation metrics in the Habitat Challenge~\cite{batra2020objectnav}, ObjectNav has attracted increasing research interest. Various approaches have been proposed to tackle this problem, including end-to-end learning~\cite{ramrakhya2023pirlnav,zhang2023layout}, modular frameworks~\cite{ChenLKGY23,chang2020semantic}, and zero-shot generalization methods~\cite{dorbala2023can,cai2024bridging}, many of which have shown promising results. In particular, the open-vocabulary setting presents a greater challenge, as it requires the agent to recognize and navigate to objects beyond a predefined set of categories. Recent advances in vision-language models (VLMs)~\cite{radford2021learning,li2023blip} and large language models (LLMs)~\cite{openai2024gpt4technicalreport,liu2024deepseek,bai2025qwen2} have opened new possibilities for solving this problem, leading to a surge of LLM-based approaches~\cite{majumdar2022zson,yokoyama2024vlfm,Zhou2023ESCEW,wu2024voronav,yu2023l3mvn,zhao2025imaginenav,nie2025wmnav}.

Despite this progress, several limitations remain in existing methods. First, many approaches rely on imitation learning~\cite{ramrakhya2022habitatweblearningembodiedobjectsearch} or reinforcement learning~\cite{wortsman2019learning}, which are often confined to narrow domains or limited object categories—making them unsuitable for real-world open-vocabulary scenarios. Second, a significant number of prior works depend on precise depth images to construct 2.5D scene representations for spatial reasoning. However, in both physical robotic systems and virtual agents, RGB-only input is far more common. Dependence on depth information thus constrains the applicability and generalizability of these methods. Third, many recent methods~\cite{yokoyama2024vlfm,long2024instructnav,nie2025wmnav} require prebuilt metric maps for navigation, which involve costly depth sensors and accurate robot localization. Such dependencies increase the system's hardware burden and reduce robustness in dynamic or noisy environments. To address these challenges, a few recent works~\cite{majumdar2022zson,cai2024bridging,zhao2025imaginenav} have proposed mapless open-vocabulary navigation frameworks. However, these methods often make decisions based only on the agent’s current observation, ignoring historical trajectory information. As a result, they are prone to local deadlocks—e.g., repeatedly revisiting already explored areas due to lack of memory.

To this end, we propose \textbf{PanoNav}, a \textbf{Pano}ramic Scene Parsing-based \textbf{Nav}igation framework for RGB-only, mapless, and open-vocabulary Object Goal Navigation. PanoNav features two core components: (1) Panoramic Scene Parsing, which leverages six multi-view RGB observations and an open-source multimodal large language model (MLLM) to build a panoramic spatial understanding without requiring depth sensors or metric maps; and (2) Dynamic Memory-Guided Decision-Making, which integrates historical trajectory information via a Dynamic Bounded Memory Queue to improve exploration efficiency and avoid local deadlocks. To the best of our knowledge, PanoNav is the first RGB-only, mapless open-vocabulary ObjectNav framework that addresses the local deadlock issue. Experiments on the HM3D dataset demonstrate that PanoNav outperforms state-of-the-art baselines such as PixNav~\cite{cai2024bridging} and ZSON~\cite{majumdar2022zson} under the same settings in terms of both Success Rate (SR) and Success weighted by Path Length (SPL), and even surpasses several map-based and closed-vocabulary methods.

In summary, our main contributions are as follows:

\begin{itemize}
    \item We propose PanoNav, an RGB-only framework that leverages multi-view visual input and vision-language reasoning to perform panoramic scene parsing for mapless, open-vocabulary object navigation.
    \item We introduce a Dynamic Memory-guided Decision-Making mechanism that incorporates a Dynamic Bounded Memory Queue to track visited areas and avoid local deadlocks.
    \item Extensive experiments demonstrate the superiority of our approach, showing significant improvements over state-of-the-art baselines in the open-vocabulary ObjectNav setting.
\end{itemize}

\section{Related Work}

\subsection{Open-Vocabulary Object Navigation}

While object navigation has seen significant progress and achieve accepted performance, many existing methods~\cite{yadav2022offlinevisualrepresentationlearning,ramrakhya2022habitatweblearningembodiedobjectsearch,ramrakhya2023pirlnav,ye2021auxiliary} remain limited to predefined object categories, failing to generalize to the open-vocabulary real-world scenarios. Therefore, open-vocabulary object navigation has gained growing attention in the research community. \citet{gadre2023cows} suggested a baseline algorithm named CoW for language-driven zero-shot object navigation by using open-vocabulary model. SSNet~\cite{zhao2023zsgn} addresses the generalization gap in ObjectNav for new categories via a semantic similarity network. VLFM~\cite{yokoyama2024vlfm} leverages vision-language models to generate semantic value maps for frontier-based exploration~\cite{topiwala2018frontierbasedexplorationautonomous} without task-specific training.

\subsection{Large-Language-Models for Embodied Navigation}

In recent years, research on foundation models has flourished, with numerous studies applying LLMs to embodied navigation tasks, including vision-language navigation (VLN), demand-driven navigation (DDN), and object navigation. NavGPT~\cite{zhou2024navgpt} is an navigation system that leverages LLMs for instruction decomposition and commonsense reasoning to achieve decision-making, ultimately enabling effective VLN. \citet{qiao2025opennav} present Open-Nav for zero-shot VLN in continuous environments using open-source LLMs, reducing costs and enhancing privacy protection. SG-Nav~\cite{yin2024sg} advances ObjectNav by representing environments as dynamically updated 3D scene graphs, where LLMs leverage hierarchical node relationships to prioritize exploration frontiers. VoroNav~\cite{yin2024sg} is the another method for ObjectNav, which leverages LLMs to reason about topological waypoints extracted from graphs. \citet{long2024instructnav} attempted to integrate navigation tasks with diverse linguistic instructions, proposing the InstructNav, which makes great use of potential of LLMs to integrate different types of instruction into its Dynamic Chain-of-Navigation module. These LLM-based approaches often rely on the metric map for navigation and decision-making.

\subsection{Mapless Object Navigation}

Recently, some progress has been made in mapless ObjectNav within the field of embodied navigation. ZSON~\cite{majumdar2022zson} achieves mapless zero-shot object navigation by leveraging the multimodal representation and alignment capabilities of CLIP~\cite{radford2021learning}, enabling generalization in open-world scenarios. ImagineNav~\cite{zhao2025imaginenav}predicts the next viewpoint using a novel synthesis model, then employs a vision-language model to select the most promising direction. This approach can achieve mapless object navigation. PixelNav~\cite{cai2024bridging} represents a zero-shot navigation approach that operates solely on RGB images, leveraging panoramic image analysis and LLMs to identify exploration-worthy pixels. However, these mapless approaches typically make decisions based solely on the current situation without considering historical information, which can easily lead to local deadlocks.

\begin{figure*}[t]
\centering
\includegraphics[width=\linewidth]{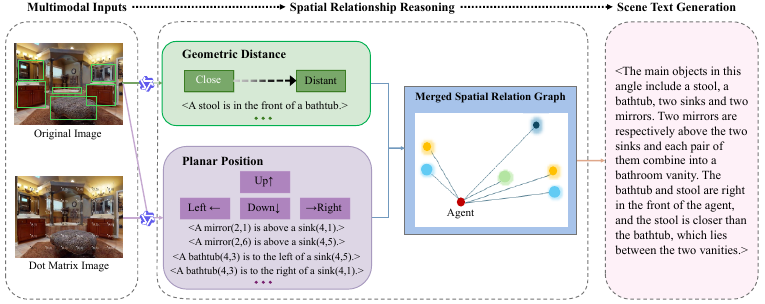}
\caption{Spatial relationship parsing in Panoramic Scene Parsing. The MLLM processes both the original RGB image and its corresponding dot matrix image to extract geometric distance and planar positional relationships between objects, producing textual descriptions of the spatial scene.
}
\label{fig:parsing}
\end{figure*}

\section{Mapless Zero-Shot Object Navigation}

\subsection{Problem Formulation}

In the Object Goal Navigation task, an agent must explore an unfamiliar indoor environment to locate an instance $I$ of a specified object category $C$ (e.g., bed, toilet, chair) based solely on a linguistic instruction. Each episode begins with the agent randomly placed in an unseen environment without any prior map or spatial knowledge.

At each timestep $t$, the agent receives an egocentric panoramic observation $\mathbf{V}_t$, consisting of six RGB images $\{\mathbf{V}_t^i\}_{i=1}^6$, captured at 60-degree intervals. No additional sensor data (e.g., depth, GPS) is provided. The agent must then choose an action to navigate toward the target. The task is considered successful if the agent stops within a predefined distance threshold of the object. The action space includes six discrete commands:
\textit{Stop, MoveAhead, TurnLeft, TurnRight, LookUp, LookDown}.
Specifically, \textit{MoveAhead} advances the agent by 0.25 meters, while \textit{TurnLeft} and \textit{TurnRight} rotate the agent by 30 degrees.

\subsection{PanoNav Framework}

PanoNav is a mapless, RGB-only navigation framework for open-vocabulary ObjectNav. Its architecture is illustrated in Fig.~\ref{fig:overview}. At each timestep, the agent captures a panoramic image set composed of six RGB views taken at 60-degree intervals. These images are first converted into corresponding dot matrix representations to enhance spatial structure cues. The panoramic RGB and dot matrix images are then fed into the Panoramic Scene Parsing module, which outputs: 1) local textual descriptions for each directional view, and 2) a global scene summary describing the agent’s current surroundings. These textual cues, along with a Dynamic Bounded Memory Queue that stores recent global summaries, are input into the Memory-Guided Decision-Making module. This module selects the most promising action based on both current and historical context. The chosen action is then executed by the motion controller.

Each component is detailed in the following sections.

\subsection{Panoramic Scene Parsing}

Recent years have witnessed increasing interest in using multimodal large language models (MLLMs) for scene understanding~\cite{bai2025qwen2,liu2023visual}. While previous zero-shot object navigation methods have shown potential, they often suffer from limited perceptual scope or rely heavily on depth images to construct spatial 2.5D scene representations. To address these challenges, we propose a Panoramic Scene Parsing module that leverages multiview RGB images and the reasoning capabilities of MLLMs to generate both fine-grained local directional parsing and a holistic global panoramic summary at each timestep.

\subsubsection{Local Directional Parsing}
The goal of local directional parsing is to extract rich contextual information from each directional view, including object presence, spatial relationships among objects, room type, the likelihood of target object presence, and overall information richness. In particular, parsing spatial relationships plays a crucial role in enhancing the agent's spatial understanding.

To fully exploit the MLLM’s capability in spatial relationship reasoning, we introduce dot matrix images as additional inputs.
At each timestep $t$, the agent captures an egocentric panoramic observation composed of six RGB views, denoted as $\mathbf{V}_t = {\mathbf{V}_t^1, \ldots, \mathbf{V}_t^6}$. For each view $\mathbf{V}_t^i$, we generate a corresponding dot matrix image $\mathbf{M}_t^i$ using the Scaffold (SCA)~\cite{lei2024scaffolding} processing method:
\begin{equation}
    \mathbf{M}_t^i = \mathrm{SCA}(\mathbf{V}_t^i), \quad i \in \{1, \ldots, 6\}.
\end{equation}

As shown in Fig.~\ref{fig:parsing}, the raw RGB image $\mathbf{V}_t^i$ captures geometric distance cues, while the paired dot matrix image $\mathbf{M}_t^i$ enhances planar position understanding. Together, these complementary inputs allow the MLLM to reason over both spatial dimensions. We construct a spatial relation graph $G_t^i$ in the latent space:
\begin{equation}
    G_t^i=\mathcal{P}(\Psi(\mathbf{V}_t^i),\Phi(\mathbf{V}_t^i,\mathbf{M}_t^i)),
\end{equation}
where $\Psi(\cdot)$ extracts geometric distance relations, $\Phi(\cdot)$ parses planar positional relationships, and $\mathcal{P}(\cdot)$ aggregates these two relational representations into a unified graph. This process yields a detailed and structured description of spatial relations for each direction, significantly enhancing the agent’s local spatial awareness.

\subsubsection{Global Panoramic Summary}
Beyond local parsing, we perform a global analysis of the robot’s spatial context to derive higher-level semantic understanding. Unlike the directional views that focus on object-level detail, the global summary emphasizes holistic environmental perception. Specifically, this involves identifying what objects are generally present in the surrounding environment and determining the type of room or scene the agent currently occupies.

This global parsing provides a form of implicit self-location awareness, offering cues about the agent’s position within the broader environment (e.g., kitchen, hallway). The output of this panoramic summarization serves as a foundational input to our decision-making module, particularly in the construction of the dynamic memory state used to guide future actions.

\begin{figure}[t]
\centering
\includegraphics[width=\linewidth]{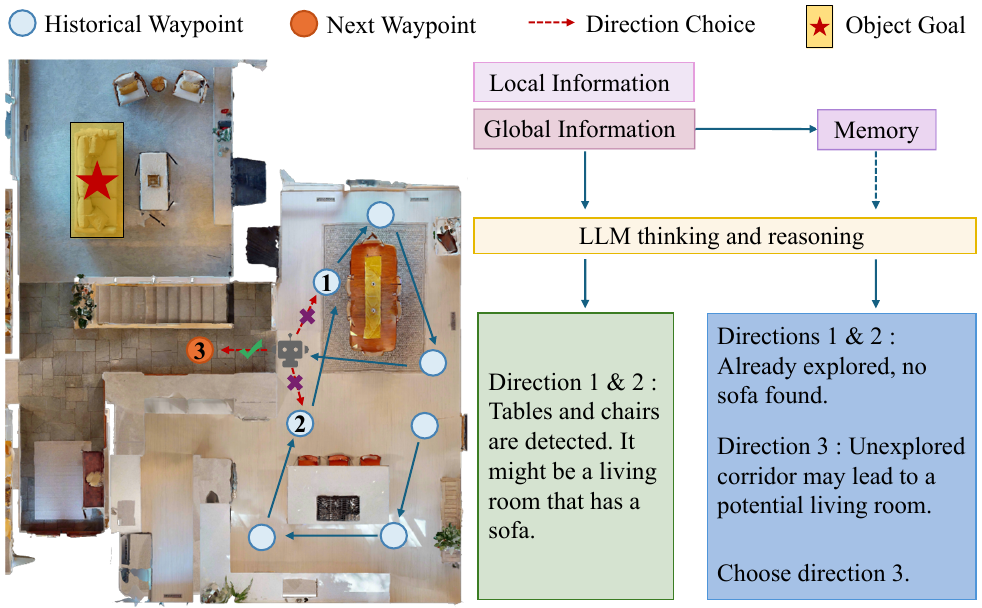}
\caption{Workflow of the Dynamic Memory-Guided Decision mechanism. Without memory, the LLM may cause the robot to revisit previously explored areas. Incorporating a memory mechanism guides the LLM to improve exploration and avoid redundant navigation.}
\label{fig:history}
\end{figure}

\subsection{Navigation Decision-Making with Dynamic Memory}

In the absence of metric maps, previous mapless zero-shot object navigation methods often fail to incorporate historical exploration information, making them susceptible to local deadlocks—situations where the agent becomes trapped in a limited area, endlessly circling within a room. We observed that agents tend to remain in regions that semantically resemble likely target locations (e.g., a living room when searching for a sofa), even if the target is not actually present. This is primarily because LLMs tend to over-rely on object-room priors during decision-making, without considering the agent’s exploration history.

As shown in Fig.~\ref{fig:history}, although the agent has already searched the right-hand region, the LLM still favors this area as the likely target location due to the contextual presence of tables and chairs—objects commonly associated with living rooms. This bias causes the agent to overlook other unexplored yet potentially relevant regions, such as corridors.
 
To alleviate this challenge, we introduce a Dynamic Memory-guided Decision-Making mechanism featuring the Dynamic Bounded Memory Queue, which effectively incorporates the robot's historical exploration data into the decision-making process. As depicted in Fig.~\ref{fig:overview}, the panoramic observation is processed through our Local-Global Panoramic Parsing module, yielding a sequence of local descriptions information $\mathbf{ld}_t^i$ and a global summary information $\mathbf{gs}_t$ for the current timestep $t$. 

We utilize the global summary description of each timestep to construct historical information, which is maintained as the Dynamic Bounded Memory Queue $\mathcal{Q}$ with a maximum length of $n$. At the initial stage of each episode, the queue is empty ($\mathcal{Q}_0 = \{~\}$). As the agent moves, new elements are continuously appended to the queue until it reaches full capacity. At this point, the queue's full flag $f_t$ is set to $1$. At timestep $t$, the queue is represented as:
\begin{equation}
    \mathcal{Q}_t=\{\mathbf{gs}_{t-n},...,\mathbf{gs}_{t-2},\mathbf{gs}_{t-1}\}.
\end{equation}
Thereafter, when a new waypoint is reached, the oldest waypoint description is dequeued while the newest one is added, maintaining a constant queue length.

\begin{table*}[h!]
  \centering
  \setlength{\tabcolsep}{6mm}{
    \begin{tabular}{cccccc}
    \toprule
    \multirow{2}[4]{*}{\textbf{Method}} & \multicolumn{3}{c}{\textbf{Navigation Setting}} & \multirow{2}[4]{*}{\textbf{SR$\uparrow$}} & \multirow{2}[4]{*}{\textbf{SPL$\uparrow$}} \\
\cmidrule{2-4}          & \textbf{Modality} & \textbf{Scope} & \textbf{Map}   &       &  \\
    \midrule
    FBE   &  RGB-D, GPS+Compass & Close-Set & map-based & 33.7  & 15.3 \\
    SemExp   & RGB-D, GPS+Compass    & Close-Set & map-based & 37.9  & 18.8  \\
    Habitat-Web & RGB-D, GPS+Compass   & Close-Set & mapless & 41.5 & 16.0 \\
    OVRL    & RGB-D, GPS+Compass   & Close-Set & mapless  & 62.0 & 26.8 \\
       VLFM    & RGB-D, GPS+Compass   & Open-Set & map-based & 52.2 & 30.4 \\
    ESC    & RGB-D, GPS+Compass    & Open-Set & map-based & 39.2  & 22.3 \\
    VoroNav  & RGB-D, GPS+Compass   & Open-Set & map-based & 42.0 & 26.0  \\
    L3MVN  & RGB-D, GPS+Compass    & Open-Set & map-based & 50.4 & 23.1 \\
    
    ImagineNav    & RGB-D    & Open-Set & mapless & 53.0 & 23.8 \\

    \midrule
    ZSON    & RGB Only    & Open-Set & mapless & 25.5 & 12.6 \\
    PixNav  & RGB Only  & Open-Set & mapless & 37.9 & 20.5 \\
    PanoNav (Ours)  & RGB Only  & Open-Set & mapless & \textbf{43.5} & \textbf{23.7} \\

    \bottomrule
    \end{tabular}%
    }
\caption{Comparison of Success Rate (SR) and Success-weighted Path Length (SPL) across state-of-the-art methods under various navigation settings. ZSON, PixNav, and our method are the only mapless, open-vocabulary approaches using RGB-only inputs. Our method outperforms both and even surpasses several methods with stronger assumptions.}
  \label{tab:comparative}%
\end{table*}%


The decision system receives local descriptions information and global summary information of the current waypoint as inputs. When the Dynamic Bounded Memory Queue is not full, only these two inputs are considered by the LLM for determining the promising direction. Once the queue reaches full capacity, the system additionally incorporates the memory queue as input for the decision-making process:
\begin{equation}
    \begin{cases}
    \mathbf{r}_t = \mathcal{F}(\mathbf{ld}_{t},\mathbf{gs}_{t}),&f_{t}=0\\
    \mathbf{r}_t = \mathcal{F}(\mathbf{ld}_{t},\mathbf{gs}_{t},\mathcal{Q}_{t}),&f_{t}=1
    \end{cases},
\end{equation}
where $\mathcal{F}(\cdot)$ donates the decision LLM, and $\mathbf{r}_t$ represents the decision result at timestep $t$, which includes both the decision direction and a flag indicating whether the target has been found. 

Once a decision is made, we employ PixNav~\cite{cai2024bridging} as the motion controller to carry out the navigation action. In summary, the Dynamic Bounded Memory Queue empowers the decision system to reason over historical exploration contexts, reducing the risk of local deadlocks and enabling more efficient object search in mapless open-vocabulary settings.

\begin{figure*}[t]
\centering
\includegraphics[width=\linewidth]{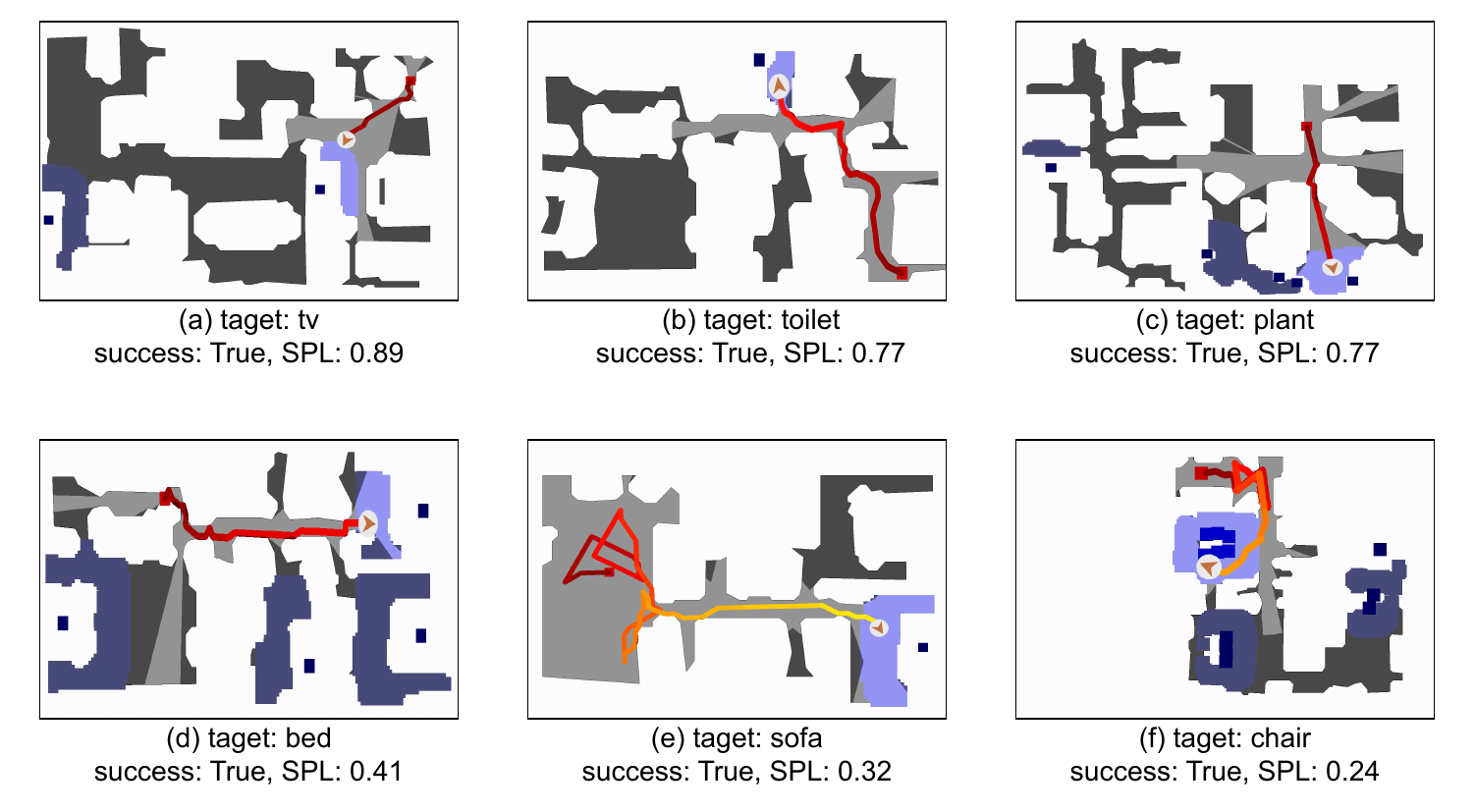}
\caption{The top-down view of the navigation trajectories includes six representative examples with different object types. Cases (a)-(d) demonstrate several smooth navigation scenarios, while cases (e) and (f) present more complex situations where the robot initially circled within a local space before successfully escaping and ultimately reaching the target.}
\label{fig:exp}
\end{figure*}

\begin{figure}[h!]
\centering
\includegraphics[width=\linewidth]{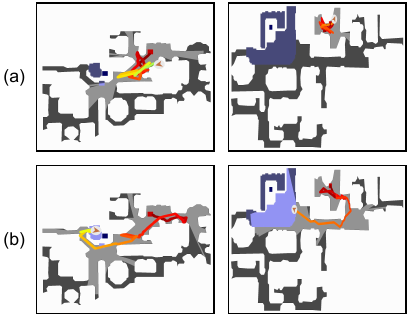}
\caption{Visualization Results of the Deadlock Avoidance Test. (a) Memory-less decision (b) Memory-guided decision.}
\label{fig:exp_memory}
\end{figure}

\section{Experiment}
\subsection{Benchmark and Evaluation Metrics}

We evaluate the effectiveness of our proposed method using the Habitat simulator on the ObjectNav benchmark, specifically the Habitat-Matterport 3D Research Dataset (HM3D)~\cite{ramakrishnan2021habitat}, a widely used benchmark for object goal navigation. HM3D provides high-fidelity reconstructions of 20 complete buildings. For our experiments, we randomly selected 200 episodes from its validation set.
In our framework, the Local-Global Panoramic Parsing module utilizes Qwen-2.5-VL~\cite{bai2025qwen2} as the Multimodal Large Language Model (MLLM), while the Memory-Guided Decision-Making System is powered by DeepSeek-V3~\cite{liu2024deepseek} as the Large Language Model (LLM).

We evaluate navigation performance using two standard metrics:
\begin{itemize}
    \item Success Rate (SR): the percentage of episodes in which the agent successfully navigates to the target object within a defined threshold.
    \item Success-weighted Path Length (SPL): a measure of navigation efficiency, computed as:
\end{itemize}
\begin{equation}
    SPL=\frac{1}{N}\sum_{i=1}^{N}s_{i}\left(\frac{\ell_{i}}{\max(	\rho_{i},\ell_{i})}\right)
\end{equation}

where $\rho_{i}$ and $\ell_{i}$ denote the agent path length and ground true path length in episode $i$, respectively. $s_{i}$ is a binary success indicator, which means the $SPL_i = 0$ when the certain episode $i$ fails.

\begin{table}[thbp]
  \centering
  \renewcommand{\arraystretch}{1.2}

    \setlength{\tabcolsep}{1.8mm}{
    \begin{tabular}{ccccc}
    \toprule
    & \textbf{SR$\uparrow$} & \textbf{SPL$\uparrow$}   
    & \textbf{DTS~(f)$\downarrow$} & \textbf{ER$\uparrow$} 
    \\
    \midrule
    without Memory & 12.0   & 4.9    & 6.7  & 32.0      \\
    with Memory & \textbf{48.0}    & \textbf{19.2}     &  \textbf{4.7} & \textbf{82.0}      \\
    \bottomrule
    \end{tabular}%
    }
  \caption{Results of deadlock avoidance test. The SR, SPL, DTS~(f), and ER correspond to Success Rate, Success-weighted Path Length, Distance to Success (in fail cases), and Escape Rate, respectively.}
  \label{tab:memory}%
\end{table}%

\subsection{Performance on Mapless Zero-Shot ObjectNav}

We selected multiple object navigation baselines with diverse navigation setting for comparative analysis. As shown in Table~\ref{tab:comparative}, these baselines were systematically categorized along three dimensions in the navigation framework: (1) Modality: the information required for the navigation method; (2) Scope: whether this method supports open-vocabulary navigation; (3) Map: whether the approach relies on building a metric map. 

For close-set object navigation approaches, we selected four baseline methods, including two map-based approaches (FBE~\cite{topiwala2018frontierbasedexplorationautonomous} and SemExp~\cite{chaplot2020objectgoalnavigationusing}) that utilize RGB-D information to construct metric maps for guiding agent exploration. Additionally, we incorporated two mapless methods: Habitat-Web~\cite{ramrakhya2022habitatweblearningembodiedobjectsearch}, which employs large-scale imitation learning by replicating human task execution strategies, and OVRL~\cite{yadav2022offlinevisualrepresentationlearning}, which obtains a powerful visual encoder through extensive pre-training.

We also introduced several open-vocabulary ObjectNav methods as baselines. Among them, VLFM~\cite{yokoyama2024vlfm}, ESC~\cite{Zhou2023ESCEW}, VoroNav~\cite{wu2024voronav}, and L3MVN~\cite{yu2023l3mvn} are map-based approaches that rely on metric maps, employing different models to extract richer semantic features to guide the LLM decision. ImagineNav~\cite{zhao2025imaginenav} represents a mapless method that imagines future observation images and selects the most promising future direction for navigation. However, it incorporates spatial knowledge by using depth information during training of its imagination model. ZSON~\cite{majumdar2022zson} and PixNav~\cite{cai2024bridging} are the two methods most similar to our approach in navigation settings, as they both require only RGB data and can achieve mapless open-vocabulary object navigation.

Table~\ref{tab:comparative} presents the comparative results of our proposed PanoNav against previous methods on the HM3D dataset, and our method achieves performance scores of 43.5\% in SR and 23.7\% in SPL. Compared with the baseline PixNav, which has the same navigation settings as ours, PanoNav demonstrates a 14.76\% improvement in SR and a 15.61\% improvement in SPL, respectively. Notably, the SR of our method even surpasses several close-set methods (FBE, SemExp, Habitat-Wen) and map-based methods (FBE, SemExp, ESC, VoroNav).

Fig.~\ref{fig:exp}  presents the navigation trajectory results for locating different types of objects. As shown in Fig.~\ref{fig:exp}~(a-d), the agent demonstrates the capability to rapidly and accurately localize targets and navigate to them in some scenarios. In more challenging environments with potential deceptive (~Fig.~\ref{fig:exp}~(e-f)~), the agent exhibits the ability to recognize if it's trapped in a deadlock situation and actively attempts to escape the current area to continue searching for the target.

\begin{table*}[tbp]
  \centering
  \renewcommand{\arraystretch}{1.2}
  \setlength{\tabcolsep}{3.5mm}{
    \begin{tabular}{ccccc}
    \toprule
    \textbf{Panorama Views} & \textbf{Decoupled Parsing and Decision} & \textbf{Memory Guidance} & \multicolumn{1}{l}{\textbf{Success Rate $\uparrow$}} & \multicolumn{1}{l}{\textbf{SPL $\uparrow$}} \\
    \midrule
    \ding{55} & \ding{51} & \ding{51} & 19.5  & 9.97 \\
    \ding{51} & \ding{55} & \ding{55} & 35.0    & 20.47 \\
    \ding{51} & \ding{51} & \ding{55} & 38.5  & 22.57 \\
    \ding{51} & \ding{51} & \ding{51} &\textbf{43.5} & \textbf{23.73}  \\
    \bottomrule
    \end{tabular}%
    }
 \caption{Results of Ablation Study.}
  \label{tab:ablation}%
\end{table*}%

\subsection{Deadlock Avoidance Test}

To further validate the effectiveness of our proposed Dynamic Bounded Memory Queue, we designed an interesting Deadlock Avoidance Test. Specifically, we selected several highly deceptive episodes where the agent's starting positions were either: (1) locate where the target object would most likely appear (but is actually absent), such as ``starting in a living room looking for a sofa'', or (2) spatial boundary areas, such as ``starting in a hallway looking for a toilet''. 

We conducted 10 repeated experiments on our selected 5 episodes, evaluating performance differences between using the memory queue and not using it. We evaluated by using four metrics. SR and SPL were the same metrics in the comparative experiments. Distance to Success (DTS) refers to the distance between the robot's stopping position and the object location. We selected failed navigation cases to calculate the mean DTS~(f). Escape Rate (ER) means the percentage of the robot escaping from a local space; we defined an escape as unsuccessful if the robot stops near its starting point.

As shown in Table~\ref{tab:memory}, both the SR and SPL metrics demonstrate that agents equipped with memory guidance exhibit significantly higher chances to find the object, achieving success rate up to four times greater than the memory-less variant. Through comparative analysis of the DTS~(f), we observed that even in unsuccessful episode, memory-guided navigation enables agents to approach object more closely. Furthermore, memory-guided method achieves an impressive escape rate of 82.0\%, substantially outperforming the memory-less version (32.0\%). The superior performance in both DTS~(f) and ER metrics shows that the dynamic memory possesses stronger exploratory capabilities while avoiding deadlock to limited regions.

Fig.~\ref{fig:exp_memory} presents two visualized navigation cases. The top row (a) shows trajectories without memory guidance, while the bottom row (b) demonstrates trajectories with memory guidance. Comparative analysis reveals that agents lacking dynamic memory exhibit repetitive looping behavior, as the LLMs may consider previously visited locations as the optimal choices.

\subsection{Ablation Study}
To evaluate the contribution of each component in our proposed framework, we conduct a series of ablation experiments.

\subsubsection{Panorama View vs Limited View} 
Our method leverages six-view panoramic observations for comprehensive scene understanding. However, collecting six views introduces additional observation steps. To assess the trade-off, we compare this setup with a limited three-view (forward-facing) configuration. As shown in the first row of Table~\ref{tab:ablation}, reducing to three views significantly degrades performance, with the success rate dropping to 19.5\% and SPL to 9.97\%. The narrower field of view limits spatial awareness, causing the agent to miss critical navigational cues or make suboptimal decisions in complex environments.

\subsubsection{Decoupled Parsing and Decision vs One-Step Decision} 

PanoNav adopts a decoupled parsing and decision strategy: an MLLM first parses RGB inputs into structured textual descriptions, which are then used by an LLM for decision-making. We compare this with a one-step approach where the MLLM directly outputs navigation decisions without intermediate descriptions. As shown in the second row of Table~\ref{tab:ablation}, the one-step method achieves moderate success but underperforms compared to the decoupled pipeline. This suggests that explicitly separating perception and reasoning helps reduce cognitive overload and encourages step-by-step processing. Current MLLMs may struggle with end-to-end reasoning when faced with rich, unstructured visual input.

\subsubsection{Memory Guidance vs No Memory Guidance} 

We further evaluate the impact of the Dynamic Bounded Memory Queue by removing it from the framework. As shown in the third row of Table~\ref{tab:ablation}, the absence of memory guidance significantly weakens navigation performance. Without memory, the agent is more prone to local deadlocks and inefficient exploration. In contrast, our memory-guided approach enables better long-horizon planning and reduces revisiting previously explored areas, which is also supported by results in the Deadlock Avoidance Test.

\section{Conclusion}

In this work, we propose a mapless open-vocabulary object navigation approach PanoNav, that requires only RGB image inputs for navigation. Facing the limitations of most current ObjectNav methods that require mapping or depth images to parse spatial structures, we propose Panoramic Scene Parsing, which unlocks the spatial relationship parsing capability of MLLMs and provides more comprehensive reference information for decision-making systems. Furthermore, to mitigate the local deadlock issue commonly encountered in mapless methods, we propose a Dynamic Memory-guided Decision-Making mechanism that enables escape from local regions demonstrates superior exploration performance. Extensive experimental results have showed the effectiveness of both our overall framework and its individual components. In future work, we will explore leveraging multimodal information to construct memory queues for more robust mapless object navigation.

\section{Acknowledgments}
This work is funded by the Young Elite Scientist Sponsorship Program by CAST (No. YESS20220181) and the National Natural Science Foundation of China (NFSC) under the Grant Number 62573370.

\bibliography{references/mybib}

\end{document}